\documentclass[runningheads]{llncs}
\usepackage[T1]{fontenc}
\usepackage{cite}
\usepackage{float}
\usepackage{bbding} 
\usepackage{amsmath,amssymb,amsfonts}
\usepackage{times}
\usepackage{latexsym}
\usepackage[table]{xcolor}
\usepackage{longtable}
\usepackage{multirow}
\usepackage{rotating}
\usepackage{enumitem}
\usepackage{placeins}
\usepackage{supertabular}
\usepackage{threeparttable}
\usepackage[vlined,ruled,linesnumbered]{algorithm2e}
\usepackage{booktabs} 
\usepackage{multirow}
\usepackage{rotating}
\usepackage{makecell}
\usepackage{subcaption}
\usepackage{placeins}
\usepackage{amsfonts}
\usepackage{algorithmic}
\usepackage{textcomp}
\usepackage{dblfloatfix} 
\usepackage{xcolor}
\usepackage{arydshln}
\usepackage{graphicx}

\begin{document}
\title{SMSD: A Semantic Misinformation Spreader Detection Approach}
\titlerunning{SMSD}
%
\author{Zeeshan Bin Siddique \Envelope\inst{1}\orcidID{0000-0002-3208-4064} \and
Michael Bewong\inst{1}\orcidID{0000-0002-5848-7451} \and
Md Geaur Rahman\inst{1}\orcidID{0000-0002-0710-3753} \and
Sajal Halder\inst{2}\orcidID{0000-0002-0965-6255} \and
Selasi Kwashie\inst{3}\orcidID{0000-0003-4014-4976}}
\authorrunning{ZB. Siddique et al.}
%
\institute{School of Computing, Mathematics and Engineering, Charles Sturt University, Australia.\\ 
\email{\{zsiddique, mbewong, grahman\}@csu.edu.au} \and
School of Computing Technologies, RMIT University, VIC, Australia.\\
\email {sajal.halder@rmit.edu.au}\\
\and Independent Researcher, Adelaide, Australia. \\
\email {selasi.kwashie@mymail.unisa.edu.au}}
%
\maketitle              
\pagenumbering{gobble}
\begin{abstract}
\label{abstract}
Misinformation spreaders on social media significantly amplify false narratives, making early detection critical to mitigate their spread. Existing approaches often overlook evolving user behavior, introduce semantic noise by ignoring user stance, and fail to handle newly emerging users effectively. To address these limitations, we propose a \underline{S}emantic \underline{M}isinformation \underline{S}preader \underline{D}etection approach (SMSD) that models misinformation propagation as a temporal, user-centric, and behavior-driven learning problem. SMSD adopts a multimodal architecture that incorporates three categories of features: user-level, network-level, and conversation propagation. A stance-aware labeling strategy is also introduced to differentiate supportive, opposing, questioning, and neutral users towards misinformation claims. Additionally, a monthly quantile analysis is introduced to simulate realistic, evolving user participation. Initial experimental results on the FibVid COVID-19 benchmark dataset demonstrate that SMSD improves recall by 45.0\%, F1 by 49.5\%, and MCC by 74.6\% under stance-aware labeling and by 17.7\%, 7.4\%, and 11.4\%, respectively, under threshold-based labeling. It consistently outperforms all baselines in both settings.

\end{abstract}
\section{Introduction}
\label{sec:Intro}
Social media platforms such as Facebook and X (formerly Twitter) serve as primary modes of communication and provide opportunities to broadcast ideas and express opinions on various issues \cite{1lu2020gcan}. This inadvertently leads to several societal challenges, such as the spread of misinformation. More than 40\% of traffic to websites containing misinformation comes from links on Facebook, Instagram, and X due to the broad reach of these social media platforms \cite{7shu2022detecting}. In 2019, the COVID-19 pandemic clearly demonstrated that the spread of misinformation can adversely affect society \cite{17banerjee2021retracted}. For example, more than 700 people died in Iran as a result of the claim that a disinfectant could treat COVID-19 \cite{18iran_methanol_covid2020}.

The spread of misinformation has two key impacts. First, constant exposure might alter one’s perception of information's veracity, especially if false information aligns with one’s belief system. Second, when false information proliferates, it becomes more difficult to distinguish fact from fiction; hence, consumers question the credibility of all information \cite{101schoenmueller2025shares}. Therefore, it is essential to identify users who willingly or inadvertently spread false information across social media.

Detecting misinformation spreaders on social media remains a complex and evolving challenge. Existing approaches, such as Ahmed et al. \cite{3aahmed2024social} and Tommasel et al. (CounterSpread) \cite{82tommasel2025countering}, achieve strong performance by integrating graph convolutional network (GCN)-based social interaction features with psycholinguistic and conversation propagation features. However, three systematic limitations remain unaddressed.

First, threshold-based labeling fails to distinguish between users who intentionally promote false claims and those who share content to debunk or criticize it. This labeling approach labels users as misinformation spreaders based exclusively on the proportion of their posts that relate to verified false claims. Users are labeled as spreaders if more than 50\% of their posts contain false claims. This introduces semantic labeling noise. The same issue persists even in stance-informed studies. For example, Soga et al.\ \cite{5asoga2024exploiting} implement stance modeling at the news classification level on the FibVID dataset, yet user labels are still derived from thresholds and retain the same ambiguity.

Second, user behavior is treated as static and temporally homogeneous. Ahmed et al. \cite{3aahmed2024social} do not include any temporal dimension in their model or evaluation. Their approach considers the entire user history at once and uses a single train/test split, making it impossible to model early detection and behavioral evolution. CounterSpread \cite{82tommasel2025countering} implements a global cumulative temporal split but still treats all users in each evaluation slot as a uniform population, thereby failing to distinguish between newly active users and long-term participants. Neither framework adequately reflects the realistic scenario in which new users continuously join misinformation discussions at various stages of a diffusion event.

Third, early detection performance remains critically low in the initial stages of information diffusion. CounterSpread \cite{82tommasel2025countering} shows limited early-stage performance, with a recall of 0.0\% in Q1 and reaching 3.8\% by Q5. This suggests limited effectiveness in identifying spreaders when interaction history is minimal. Plepi et al. \cite{6aplepi2022temporal} partially address this limitation by constructing temporal graphs; however, their method still requires content similarity signals to accumulate across multiple time frames before reliable predictions can be made. Neither framework is designed for nor evaluated on the zero-history scenario involving genuinely new users who have recently joined a discussion and have limited interaction history.

To address these limitations, we propose SMSD, a semantic misinformation spreader detection approach with four contributions. 
\vspace{-0.15cm}
\begin{itemize}
    \item First, we introduce stance-aware labeling to distinguish users who support misinformation from those who question or oppose it, reducing semantic labeling noise.
    \item Second, we implement a user-centric temporal split by partitioning each calendar month independently, which better models the emergence of new users compared to global quantile splits.
    \item Third, we incorporate behavioral activity and Behavioral Language for Online Characterization (BLOC) features to capture temporal behavioral irregularities, enabling effective early detection even with limited user interaction history.
    \item Fourth, we conduct initial experiments on the FibVid COVID-19 dataset and demonstrate that SMSD consistently outperforms baseline methods across multiple labeling strategies and temporal configurations. 
\end{itemize}
\vspace{-0.5cm}
\section{Related Work}
\label{sec:relwork}
Existing research on detecting misinformation spreaders employs a range of features, including linguistic, user-level, graph-based, behavioral, engagement, and platform-specific features.
The detection of false content has been extensively addressed in the literature \cite{133condran2025mapx}. Early studies primarily relied on linguistic characteristics \cite{1lu2020gcan, 127giachanou2022impact}. However, since fake news is intentionally crafted to mislead, distinguishing it based solely on textual analysis remains challenging. Consequently, subsequent approaches have incorporated social context information, including user profiles and network features, \cite{107pillai2025breaking}. More recent developments involve applying convolutional and recurrent neural networks to learn temporal representations from textual features and, in some cases, from content propagation networks \cite{5asoga2024exploiting}.

Several methodologies have been developed to identify fake news spreaders. Verdolotti et al. \cite{103verdolotti2025predicting} and Shrestha et al.\cite{57shrestha2022characterizing} examined hand-crafted user profile features, including screen name length, user description length, number of followers and friends, account age, average number of daily statuses, and the overall sentiment score of all shared tweets. Hashemi et al. \cite{125hashemi2026enhancing} and Fawzi et al. \cite{126fawzi2024pinocchio} proposed content-based features, including word embeddings, Linguistic Inquiry and Word Count (LIWC) categories, personality traits, communication style, sentiment, emotion, and readability scores. 

Similarly, other approaches, such as Cardaioli et al. \cite{51cardaioli2021sa}, integrate user features with an average word2Vec representation of the content to detect rumor spreaders. Pillai et al. \cite{104pillai2025misinformation} considered demographic and behavioral features. Demographic features included age, gender, and political orientation, which were inferred from users' content. Behavioral features were linked to the specific time of day and day of the week when posts were shared. Ghanem et al. \cite{129ghanem2020factweet} and Tommasel et al. \cite{82tommasel2025countering} divided user timelines into fixed-size segments, thereby enabling a single user to be represented by multiple segments. These segments were described using a combination of GloVE embeddings and semantic and stylistic features, including emotions, morality, and style. Khaund et al. \cite{61khaund2021social} have demonstrated that lexical and sentiment features are most relevant for classification. In all cases, either traditional machine learning models or simple neural network models were utilized.

Other methods have focused on network topology. Kaliyar et al. \cite{73kaliyar2021deepfake} used different variants of the HITS centrality metric to rank spreaders across news-sharing networks. Rath et al. \cite{124rath2022fake} developed an inductive representation-learning framework that integrates community health and interpersonal trust to identify spreaders within tightly connected communities.

In contrast to previous methods, SMSD addresses three primary limitations of existing approaches. It replaces threshold-based labeling with a stance-aware strategy to reduce semantic noise. It employs a user-centric monthly quantile split to more accurately capture changes in user participation over time. Additionally, it incorporates BLOC and behavioral features to facilitate reliable early detection.
\section{Problem Formulation}
\label{Sec:Problemform}
\begin{figure}[t!]
    \centering
    ~\subfloat[]{ \includegraphics[width=0.45\linewidth]{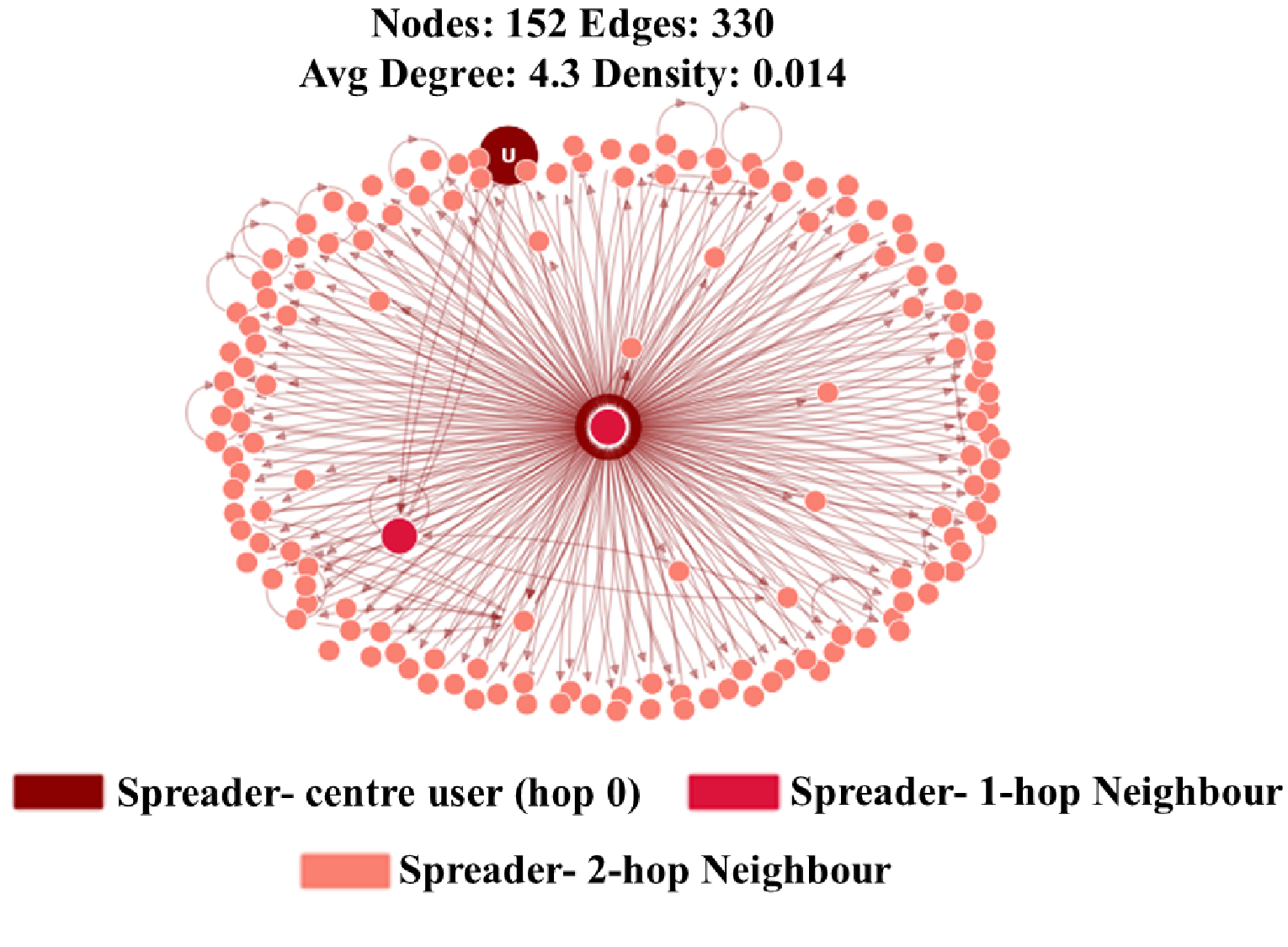}}
    ~\subfloat[]{ \includegraphics[width=0.40\linewidth]{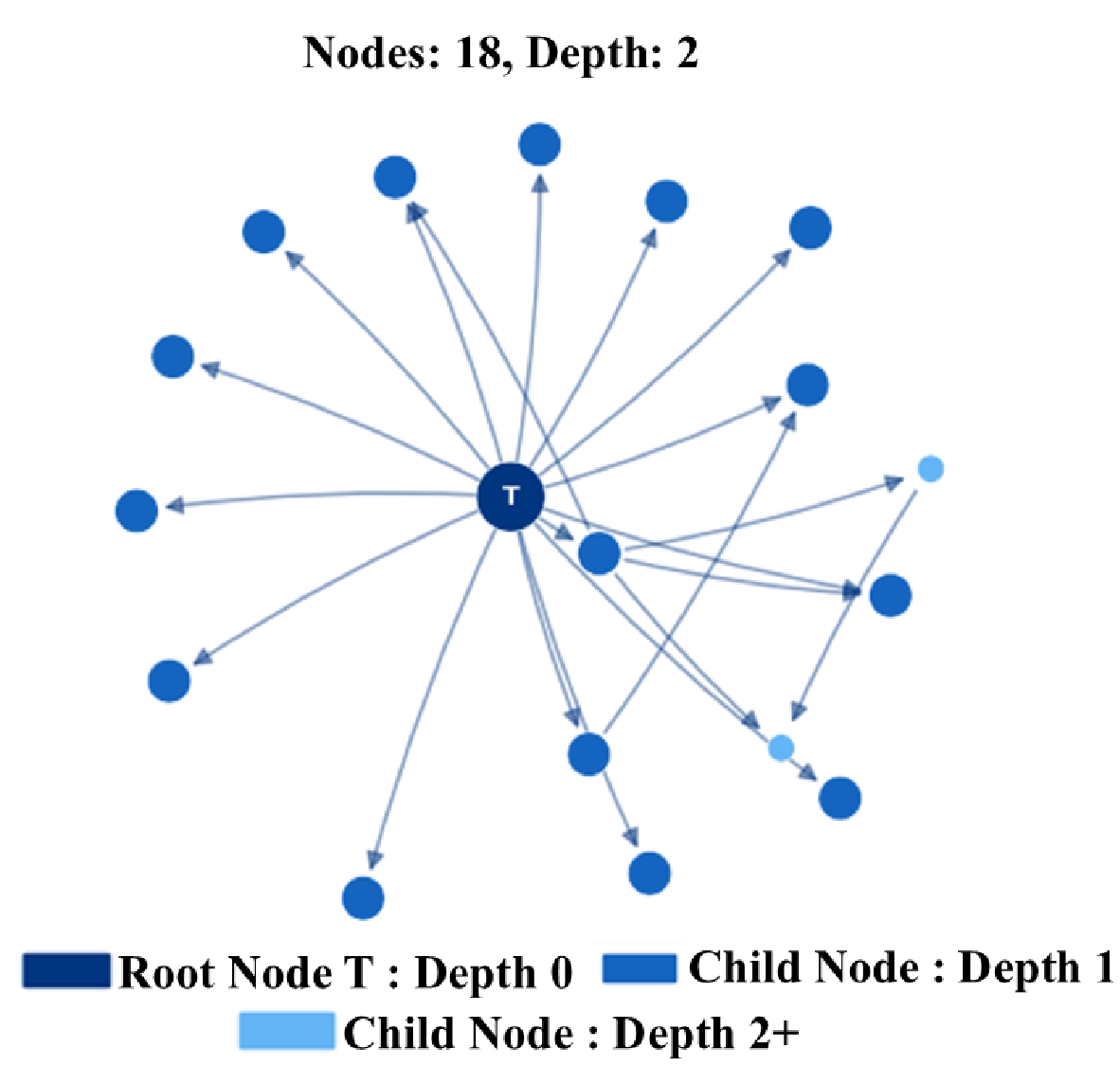}}
    \caption{(a) Social interaction subgraph $\tau(v,\mathcal{P})$, (b) Structure of a conversation propagation tree $T(v,c)$} 
    \label{Fig:Subgraphconv}
\end{figure}

We present some preliminary definitions:

\begin{definition}[Social Interaction Network]
\label{def:sin}
Given a social media platform $\mathcal{P}$ and user $v$, let $\tau(v,\mathcal{P})$ be a function that returns a subgraph $(V,E)$ of $v$'s interactions with others, such that each node $v_i \in V$ is a user, and each edge $(v,v_i) \in E$ indicates an interaction, e.g. a reply or mention, retweet, quote, or follower-followee relationship.

\end{definition}
This network captures the structural and relational characteristics of $v$ within the broader social ecosystem. Figure \ref{Fig:Subgraphconv} (a) shows the 3-hop social interaction network graph structure. 
\begin{definition}[Social Media Content]
\label{def:smc}
Given a user $v$ on a social media platform $\mathcal{P}$, and a timestamp $ts$, let $C(v,\mathcal{P},ts)$ be a function that returns all the content shared by the user $v$ on the platform $\mathcal{P}$ up to the point $ts$.

\end{definition}
From $C(v,\mathcal{P},ts)$ (simply denoted $C$), content-based features can be extracted to characterize the user's information content and style up to timestamp $ts$. These features include linguistic patterns, sentiment, readability, personality traits, punctuation style, and LIWC \footnote{LIWC (Linguistic Inquiry and Word Count) is a dictionary-based psycholinguistic tool that categorizes words into psychologically meaningful groups (e.g., emotions, cognition, social processes, and linguistic styles) \cite{17atausczik2010psychological}}. 

\begin{definition}[Conversation Propagation Tree]
\label{def:cpt}
Given any content $c \in C$, let $T(v,c)$ be a function that returns the conversation propagation tree for the user $v$ w.r.t. $c$, which contains the evolution and spread $c$ within the social network. 
\end{definition}
The root node of a conversation propagation tree corresponds to the source content, while child nodes represent replies, retweets, quotes, or other interactions originating from preceding nodes. This propagation tree characterizes diffusion dynamics, structural positions, and information flow associated with user participation in the dissemination of misinformation. Figure \ref{Fig:Subgraphconv} (b) represents the structure of three different conversation propagation trees. 

The problem of identifying spreaders within a network has often been defined in the literature \cite{82tommasel2025countering} as identifying users who amplify and propagate false information. Formally, it is defined as follows.

\begin{definition}[Spreader]
\label{def:spreader}
Given a set of users  $V$, a user $v \in V$ is said to be a spreader if $x\%$ of \emph{posts} made by $v$ are considered false, where $x\%$ is greater than a pre-specified threshold.   
\end{definition}

This definition has three key challenges as follows:
First, it is \emph{post-hoc} and can only be determined well after the false information has been propagated. Second, it is impossible to determine whether a new user is a spreader without historical context. Third, it is user-centric, whereas in practice the problem is item-centric. For example, in practical applications, one will be interested in answering the question \emph{``for this false social media post, who are the spreaders associated with it?" } instead of \emph{` "For this set of users, which of them are spreaders? "} 

An additional limitation of threshold-based labeling is semantic noise. For instance, users who share misinformation to debunk or criticize it receive the same spreader label as those who support it. To resolve this, we introduce a stance-aware labeling function $\phi(v,c_f)$ that aggregates tweet-level stance predictions (support, deny, query, neutral) across all of $v$'s interactions with $c_f$. Formally, a user is labeled as a spreader under the stance-aware labeling strategy if their support ratio exceeds a predefined threshold $\theta$, here $\theta$ = 0.5:
\begin{equation}
     y_i =
  \begin{cases}
    1,  & \text{if } \phi(v_i, c_f) > \theta,\\
    0, & \text{otherwise.}
  \end{cases}
  \label{eq:stance_label}
\end{equation}

This labeling function serves as a preprocessing step to construct supervision labels $y_i$.
We formulate the spreader detection problem as a binary classification task.

\begin{definition}[Problem Definition]
\label{def:problem}
Given any false content $c_f \in C$, for all users  $v\in V$ on a social media platform $\mathcal{P}$, the goal is to learn a binary function. 
\begin{equation}
 F( v , c_f, \mathcal{P} ) \rightarrow \{0,1\}, 
\label{eq:label}   
\end{equation}

where 1 defines $v$ as a spreader of the false content and 0 otherwise.
\end{definition}
For each user $v_i$, we construct user-level features, network-level features, and conversation propagation tree features.

\section{Methodology}
\label{sec:methodology}
\begin{figure}[t]
\includegraphics[width=0.99\linewidth]{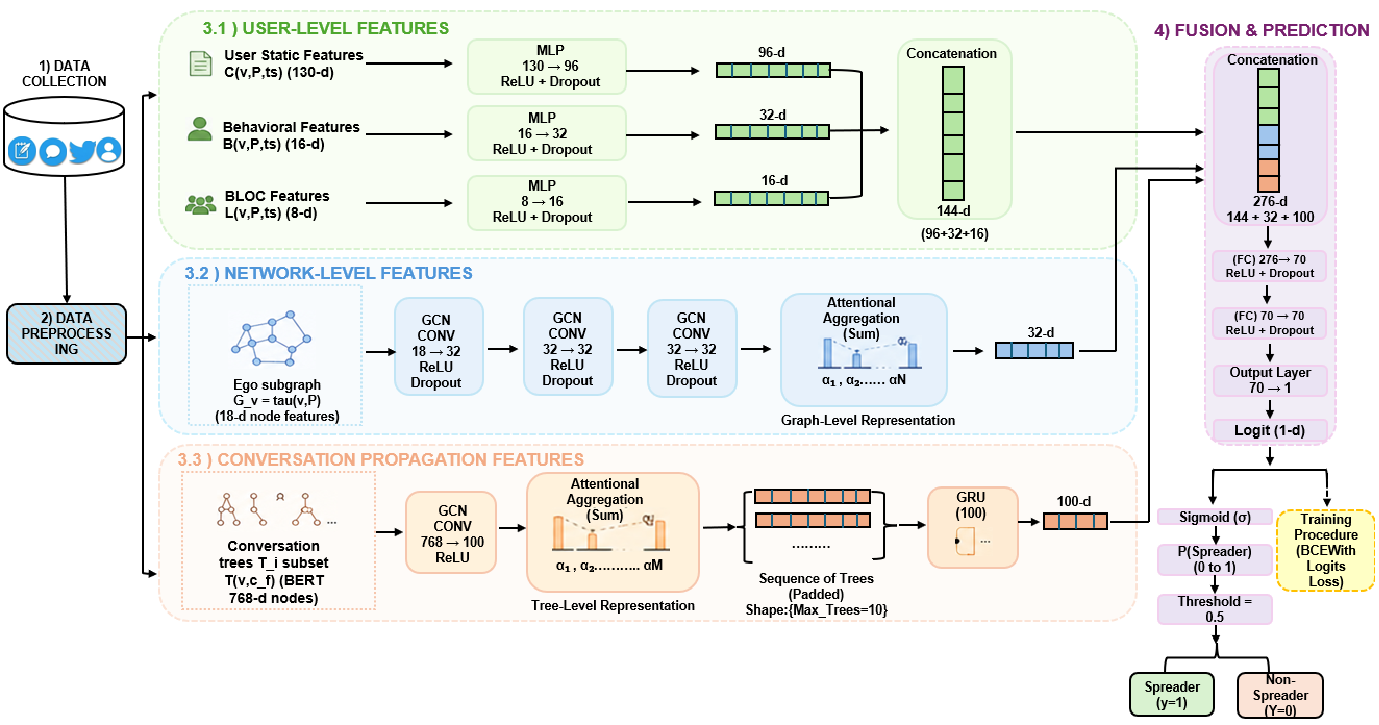}
\caption {SMSD architecture: User-level features $C,B,L$ (130/16/8-d) are encoded by parallel MLPs into $z^{user}_v$ (144-d); network-level features $\tau$ (18-d) from a 3-hop subgraph are encoded using GCN and attention pooling into $z^{graph}_v$ (32-d); conversation features $T$ (768-d BERT) are processed by GCN and GRU across up to 10 trees into $z^{conv}_v$ (100-d). These are concatenated (276-d) and passed through two 70-unit dense layers to predict $\hat{y}_v$.}
\label{Fig:Architecture}
\end{figure}

We propose an SMSD approach comprising three parallel encoding modules, followed by a fusion \& prediction module, as illustrated in Figure \ref{Fig:Architecture}. 
A user's involvement in spreading false content depends on several factors that no single data stream can fully capture. These include their linguistic style and profile, activity patterns and temporal irregularities, social network, and specific engagement with a false content item. SMSD addresses all four dimensions by constructing five feature representations for each user $v$ in relation to false content $c_f$, as illustrated in Table \ref{tab:features}.

\subsection{Data Collection \& Preprocessing}
\label{subsec:datapre}
The FibVID COVID-19 dataset \cite{3kim2021fibvid} was collected from Twitter using the platform's academic API. It contains fact-checked claims obtained from PolitiFact and Snopes, along with all related tweet conversations. Preprocessing of raw tweet data followed the pipeline described by Ahmed et al. \cite{3aahmed2024social}, which involved tweet deduplication, normalization of URLs and mentions, removal of retweet markers, and tokenization. User timelines were reconstructed in chronological order and filtered to include only users within the Largest Connected Component (LCC) of the interaction graph. 
\begin{table}[t]
\caption{Feature categories used in SMSD}
\label{tab:features}
\centering
\scriptsize
\renewcommand{\arraystretch}{1.00}

\begin{tabular}{|p{0.15\textwidth}|p{0.15\textwidth}|p{0.7\textwidth}|}
\hline
\textbf{Symbol} & \textbf{Category} & \textbf{Representative Features} \\
\hline

\multicolumn{3}{|c|}{\textbf{Module 1 -- User-Level Features}} \\
\hline

$C(v,\mathcal{P},ts)$ &
Profile, LIWC, Personality, Punctuation, Readability
&
\textbf{Profile:} follower/followee ratio, verified status, account age, profile description, screen-name statistics;
\textbf{LIWC:} social, emotional, cognitive, perceptual, health, work, religion, family, pronoun, and negation indicators;
\textbf{Personality:} Big Five and personality-disorder traits;
\textbf{Writing Style:} punctuation statistics and readability metrics (Flesch, SMOG, ARI, Dale-Chall) (130 dimensions; fused into a 96-dimensional learned vector).
\\
\hline

$B(v,\mathcal{P},ts)$ &
Behavioral Activity
&
Posting frequency, active days, repost ratio, unique hashtags, unique mentions, sentiment score, retweet activity, and temporal posting-gap statistics (16 dimensions; fused into a 32-dimensional learned vector).
\\
\hline

$L(v,\mathcal{P},ts)$ &
BLOC
&
Burst intensity, automation score, conversation ratio, pause score, URL intensity, hashtag burstiness, and action variance (8 dimensions; fused into a 16-dimensional learned vector).
\\
\hline

\multicolumn{3}{|c|}{\textbf{Module 2 -- Network-Level Features}} \\
\hline

$\tau(v,\mathcal{P})$
&
Graph / Network
&
Degree, clustering coefficient, reciprocity, PageRank, average neighbor degree, semantic similarity to neighbors, temporal reply delay, and interaction statistics (18 dimensions; encoded into a 32-dimensional GCN embedding).
\\
\hline

\multicolumn{3}{|c|}{\textbf{Module 3 -- Conversation Propagation Features}} \\
\hline

$T(v,c_f)$
&
BERT Content
&
768-dimensional tweet embeddings extracted from conversation-tree nodes and encoded using GCN and GRU layers into a 100-dimensional propagation representation.
\\
\hline

$T(v,c_f)$
&
Propagation Structure
&
Tree height, cascade size, branching factor, root distance, semantic similarity between nodes, neighborhood similarity, and interaction-count statistics.
\\
\hline

\end{tabular}
\end{table}
\subsection{User-Level Features}
\label{subsec:userlevelfeat}
In the first module, there are three groups of user-level features derived from $C(v,\mathcal{P},ts)$, $B(v,\mathcal{P},ts)$, and $L(v,\mathcal{P},ts)$, respectively. Each feature group is encoded separately using multi-layer perceptrons (MLPs) with ReLU activation and dropout, thereby creating a unified user representation. Multiple MLPs capture distinct patterns for each feature type because of their diverse characteristics. Details of all features are in Table~\ref{tab:features}.

The first group, user profile and psycholinguistic features, is extracted from $C(v,\mathcal{P},ts)$ and consists of a 130-dimensional static vector that combines profile metadata with linguistic content attributes. The second group, behavioral activity features $B(v,\mathcal{P},ts)$, comprises 16 attributes that describe user activity patterns, such as engagement and content sharing. The third group, BLOC features $L(v,\mathcal{P},ts)$, contains 8 attributes that measure temporal irregularities in user behavior. Collectively, these groups provide a comprehensive view of user profiles, communication styles, behavioral patterns, and temporal dynamics.

These features are transformed into a 96-32-16-dimensional learned
latent vector.  Letting $\mathbf{z}^{\text{static}}_v$,
$\mathbf{z}^{\text{beh}}_v$, and $\mathbf{z}^{\text{bloc}}_v$ denote
the embeddings produced from $C(v,\mathcal{P},ts)$,
$B(v,\mathcal{P},ts)$, and $L(v,\mathcal{P},ts)$ respectively, the
fused user representation is:

\begin{equation}
  \mathbf{z}^{\text{user}}_v \;\leftarrow\;
  \text{Concat}\!\left(
    \mathbf{z}^{\text{static}}_v,\;
    \mathbf{z}^{\text{beh}}_v,\;
    \mathbf{z}^{\text{bloc}}_v
  \right),
  \label{eq:user_embed}
\end{equation}

where $\mathbf{z}^{\text{static}}_v \in \mathbb{R}^{96}$,
$\mathbf{z}^{\text{beh}}_v \in \mathbb{R}^{32}$, and
$\mathbf{z}^{\text{bloc}}_v \in \mathbb{R}^{16}$, yielding
$\mathbf{z}^{\text{user}}_v \in \mathbb{R}^{144}$.
 
\subsection{Network-Level Features}
\label{subsec:network_features}
In the second module, a global social interaction graph $G$ is constructed to operationalize the social interaction network $\tau(v,\mathcal{P})$. The edges within $G$ are
determined by retweet, reply, and mention relationships among users, which are extracted from FibVid tweet metadata. For each user $v$, a 3-hop ego subgraph $G_v \subset G$ that represents $\tau(v,\mathcal{P})$, the local social neighborhood of $v$ within the broader diffusion network, is extracted. Each node is represented by an 18-dimensional feature vector comprising network- and neighborhood-based attributes.
 These subgraphs are processed with a three-layer Graph Convolutional Network (GCN) \cite{28berg2017graph}, which updates node representations as follows:
\begin{equation}
  \mathbf{H}^{(l+1)} = \sigma\!\left(
    \tilde{\mathbf{D}}^{-\frac{1}{2}}
    \tilde{\mathbf{A}}
    \tilde{\mathbf{D}}^{-\frac{1}{2}}
    \mathbf{H}^{(l)}\mathbf{W}^{(l)}
    + \mathbf{b}^{(l)}
  \right).
  \label{eq:gcn}
\end{equation}
An attention-based aggregation mechanism generates a fixed-size graph
representation:
\begin{equation}
  \mathbf{z}^{\text{graph}}_v =
  \sum_{i=1}^{N} \alpha_i \mathbf{h}_i,
  \qquad
  \alpha_i =
  \frac{\exp\!\left(g(\mathbf{h}_i)\right)}
       {\sum_{j=1}^{N} \exp\!\left(g(\mathbf{h}_j)\right)},
  \label{eq:attn_graph}
\end{equation}
where $\mathbf{h}_i$ is the embedding of node $i$ and $g(\cdot)$ is a
learnable scoring function.  The attention pooling layer produces
$\mathbf{z}^{\text{graph}}_v \in \mathbb{R}^{32}$, which serves as
the final network embedding for user $v$.

 \subsection{Conversation Propagation Features}
\label{subsec:conv_features}
In the third module, conversations are represented as tree-structured cascades. Each tree $\mathcal{T}_i$ is an instance of the conversation propagation tree $T(v,c)$. It is built using the reply and retweet relationships of a specific tweet $i$ that the user $v$ interacts with, and node features are initialized with BERT embeddings \cite{27devlin2019bert}. Users participating in multiple conversations contribute to $K{=}10$ padded trees.

Each conversation tree $\mathcal{T}_i$ is processed by a GCN to extract both structural and semantic features. An attention-based aggregation method then creates a fixed-size embedding for each tree.
\begin{equation}
  \mathbf{z}_i =
  \text{AttnPool}\!\left(
    \text{GCN}\!\left(\mathcal{T}_i,\,
    \mathbf{X}^{\text{BERT}}_i\right)
  \right).
  \label{eq:tree_embed}
\end{equation}
Since users often engage in multiple conversations, the sequence of tree embeddings is arranged in chronological order and processed using a Gated Recurrent Unit (GRU) to capture the temporal progression of user behavior:
\begin{equation}
  \mathbf{z}^{\text{conv}}_v =
  \mathbf{W}_c\!\left(
    \text{GRU}(\mathbf{z}_1, \ldots, \mathbf{z}_K)
  \right) + \mathbf{b}_c ,
  \label{eq:gru}
\end{equation}
where $\mathcal{T}_i$ denotes the $i$-th conversation tree
(an instance of $T(v,c)$), $\mathbf{X}^{\text{BERT}}_i$ are the
BERT-based node embeddings, and $\mathbf{z}^{\text{conv}}_v \in
\mathbb{R}^{100}$ summarises $v$'s conversation behavior over time.

\subsection{Fusion and Prediction}
\label{subsec:fusionpred}
After encoding all five feature representations using the three modules described above, we can now implement the binary classification function $F(\cdot)$.
Letting $\Phi(v,\mathcal{P},ts) = \{C(v,\mathcal{P},ts),\,
B(v,\mathcal{P},ts),\, L(v,\mathcal{P},ts)\}$ denote the user-level
feature set, the goal is to learn:
\begin{equation}
  F\!\Bigl(
    \tau(v,\mathcal{P}),\;
    \Phi(v,\mathcal{P},ts),\;
    T(v,c_f)
  \Bigr)
  \;\rightarrow\; \{0, 1\},
  \label{eq:Ffull}
\end{equation}
where the output label $y_v \in \{0,1\}$ is as defined in
Equation \eqref{eq:label}.
This is implemented by concatenating the latent representations
from all three branches:
\begin{equation}
  \mathbf{z}_v \;\leftarrow\;
  \text{Concat}\!\left(
    \mathbf{z}^{\text{user}}_v,\;
    \mathbf{z}^{\text{graph}}_v,\;
    \mathbf{z}^{\text{conv}}_v
  \right),
  \label{eq:fusion}
\end{equation}
where $\mathbf{z}^{\text{user}}_v \in \mathbb{R}^{144}$,
$\mathbf{z}^{\text{graph}}_v \in \mathbb{R}^{32}$, and
$\mathbf{z}^{\text{conv}}_v \in \mathbb{R}^{100}$ are concatenated
to form $\mathbf{z}_v \in \mathbb{R}^{276}$.
This combined representation is subsequently processed by two fully connected layers, each comprising 70 units with ReLU activation and dropout regularisation, prior to binary classification:
\begin{gather}
\mathbf{h}_1 = \text{Dropout}(\phi(\mathbf{W}_1\mathbf{z}_v+\mathbf{b}_1)), \quad
\mathbf{h}_2 = \text{Dropout}(\phi(\mathbf{W}_2\mathbf{h}_1+\mathbf{b}_2)) \label{eq:fc1},\\
y_v^{\text{logit}} = \mathbf{W}_3\mathbf{h}_2+\mathbf{b}_3,
\label{eq:logit}\\
\hat y_v = \sigma(y_v^{\text{logit}}) \label{eq:sigmoid},
\end{gather}
where $\phi(\cdot)$ is the ReLU activation function,
$\mathbf{W}_1, \mathbf{W}_2, \mathbf{W}_3, \mathbf{b}_1, \mathbf{b}_2,
\mathbf{b}_3$ are learnable parameters, $y^{\text{logit}}_v \in
\mathbb{R}$ is the model logit, and $\hat{y}_v \in [0,1]$ is the
predicted probability that user $v$ is a spreader.  This realizes the
binary classification function $F(\cdot)$.

\subsection{Model Training}
\label{subsec:modeltrain}
The model uses the BCEWithLogitsLoss function for training. This function combines a sigmoid activation function with binary cross-entropy, making it stable and effective for binary classification \cite{132paszke2019pytorch}. The loss function is defined as follows:
\begin{equation}
\small
\begin{aligned}
\mathcal{L} =
-\frac{1}{N}
\sum_{i=1}^{N}
w_{y_i}
\Big[
& y_i \log\left(\sigma(y_i^{\text{logit}})\right) 
& + (1-y_i)\log\left(1-\sigma(y_i^{\text{logit}})\right)
\Big] .
\end{aligned}
\end{equation}
In this context, $y_i^{\text{logit}}$ represents the predicted logit score for sample $i$, $y_i \in \{0,1\}$ represents the ground truth label, $w_{y_i}$ indicates the class weight applied to handle class imbalance, $\sigma(\cdot)$ refers to the sigmoid activation function, and $N$ is the total number of training samples.


\begin{algorithm}[b!]
\caption{SMSD: Data Pipeline and Model Architecture.}
\label{alg:smsd_arch}
\scriptsize 
\SetKwFunction{This}{this}
\KwData{ Raw tweet corpus and user set from FibVID; false content set $C_f$; fact-checking labels from PolitiFact and Snopes}

\KwResult{User representation $\mathbf{z}_v$ and prediction logit $y_v^{\text{logit}}$ for each user $v \in U$}

\textbf{Step 1: Data Collection}

Retrieve FibVID dataset: claims $C_f$, tweet corpus $\mathcal{T}$,
user set $U$. 
\hfill see Section \ref{subsec:datapre}

 \vspace{0.1cm}
\textbf{Step 2: Data Preprocessing}
\hfill 

Preprocess tweets, reconstruct user timelines, and build interaction graph G \\
Assign stance-aware or threshold-based labels yv;
\hfill see Section \ref{subsec:datapre}

\vspace{0.1cm}
\textbf{Step 3: Encode user-level features}

\ForEach{user $v \in U$}{
  Extract static features from $C(v,\mathcal{P},ts)$\\
  $\mathbf{z}^{\text{static}}_v
    \leftarrow \text{MLP}_a(\mathbf{x}^{\text{static}}_v)$
  \hfill 

  Extract behavioral features from $B(v,\mathcal{P},ts)$\\
  $\mathbf{z}^{\text{beh}}_v
    \leftarrow \text{MLP}_b(\mathbf{x}^{\text{beh}}_v)$
  \hfill 

  Extract BLOC features from $L(v,\mathcal{P},ts)$\\
  $\mathbf{z}^{\text{bloc}}_v
    \leftarrow \text{MLP}_c(\mathbf{x}^{\text{bloc}}_v)$
  \hfill 

  Fuse all user embeddings (Eq.~\eqref{eq:user_embed})\\
  $\mathbf{z}^{\text{user}}_v
    \leftarrow
    \text{Concat}(
      \mathbf{z}^{\text{static}}_v,\,
      \mathbf{z}^{\text{beh}}_v,\,
      \mathbf{z}^{\text{bloc}}_v)$
  \hfill 
}

\vspace{0.1cm}
\textbf{Step 4: Learn network representation}

\ForEach{ego subgraph $G_v \subset G$ encoding $\tau(v,\mathcal{P})$}{
  Apply three-layer GCN (Eq.~\eqref{eq:gcn})\\
  $\mathbf{H}_v
    \leftarrow \text{GCN}^{(3)}(G_v)$
  \hfill

  Generate graph embedding via attention pooling
  (Eq.~\eqref{eq:attn_graph})\\
  $\mathbf{z}^{\text{graph}}_v
    \leftarrow \text{Attention}(\mathbf{H}_v)$
  \hfill 
}

\vspace{0.1cm}
\textbf{Step 5: Encode conversation trees}

\ForEach{tree $\mathcal{T}_i \in T(v,c_f)$, up to $K=10$ trees per user}{
  Initialize tweet nodes using BERT
  embeddings
  \hfill 

  Apply GCN over tree structure (Eq.~\eqref{eq:tree_embed})\\
  $\mathbf{H}_i
    \leftarrow \text{GCN}(\mathcal{T}_i,\,\mathbf{X}^{\text{BERT}}_i)$
  \hfill 

  Generate tree embedding \\
  $\mathbf{z}_i
    \leftarrow \text{Attention}(\mathbf{H}_i)$
  \hfill 
}

\vspace{0.1cm}
\textbf{Step 6: Model temporal conversations}

\ForEach{user $v \in U$}{
  Arrange trees $\{\mathbf{z}_1,\ldots,\mathbf{z}_K\}$ chronologically;
  pad to $K\!=\!10$ if fewer trees exist

  Learn sequential dependencies (Eq.~\eqref{eq:gru})\\
  $\mathbf{z}^{\text{conv}}_v
    \leftarrow \text{GRU}(\mathbf{z}_1,\ldots,\mathbf{z}_K)$
  \hfill
}

\vspace{0.1cm}
\textbf{Step 7: Fuse and classify}

\ForEach{user $v \in U$}{
  Fuse all branches (Eq.~\eqref{eq:fusion})\\
  $\mathbf{z}_v
    \leftarrow
    \text{Concat}(
      \mathbf{z}^{\text{user}}_v,\,
      \mathbf{z}^{\text{graph}}_v,\,
      \mathbf{z}^{\text{conv}}_v)$
  \hfill 

  Generate logit via two fully-connected layers
  (Eqs.~\eqref{eq:fc1}--\eqref{eq:logit})\\
  $y^{\text{logit}}_v \leftarrow f(\mathbf{z}_v)$
  \hfill 
}

\end{algorithm}

This weighting is implemented within the BCEWithLogitsLoss function. Model optimization is performed using Adam 
(lr $= 3{\times}10^{-4}$). Training uses mini-batches of size 10 and runs for up to 20 epochs with early stopping. Early stopping is applied with a patience of 5 epochs, using the validation Matthews correlation coefficient (MCC) as the criterion. The optimal checkpoint $\mathcal{M}^{*}$ is selected for inference.
Predictions are generated by thresholding $\hat{y}_v = \sigma(y^{\text{logit}}_v)$ at 0.5.

\section{Experiments \& Results}
\label{sec:expresults}
The effectiveness of the proposed SMSD approach is evaluated through a comprehensive set of experiments.

\paragraph{\textbf{Dataset:}} The FibVid COVID-19 \cite{3kim2021fibvid} misinformation dataset is utilized. It consists of fact-checked claims and their associated Twitter conversations gathered from PolitiFact and Snopes. There are 772 such claims in the dataset, along with a vast repository of tweets discussing them. Similar to the labeling scheme used in previous research \cite{82tommasel2025countering}, we label claims as factual or misleading based on annotations from fact-checking sites. A summary of the dataset statistics is provided in Table~\ref{tab:dataset_summary}.

\begin{table}[t]
\setlength{\belowcaptionskip}{0pt}
\setlength{\abovecaptionskip}{0pt}
\caption{Summary of the FibVid COVID-19 Dataset}
\label{tab:dataset_summary}
\centering
\small
\renewcommand{\arraystretch}{1.0}
\setlength{\tabcolsep}{4pt}
\resizebox{\textwidth}{!}{
\begin{tabular}
{|p{3.7cm}|p{2.9cm}|p{3cm}|p{2.9cm}|}
\hline
\textbf{Item} & \textbf{Value} & \textbf{Item} & \textbf{Value} \\
\hline
Total users (LCC) & 24,340 &
Total tweets & 296,804 \\
\hline
Tweets from LCC users & 112,433 &
Graph nodes & 24,340 \\
\hline
Graph edges & 98,804 &
Parent--child relations & 58,694 \\
\hline
Tweet representation & BERT (768-d) &
Graph representation & 3-hop Subgraph \\
\hline
Conversation representation & Tree cascades &
Total features & 172 \\
\hline
Threshold-based labels &
\begin{tabular}[c]{@{}l@{}}
Spreaders: 8,420 \\
Non-Spreaders: 15,920
\end{tabular}
&
Stance-aware labels &
\begin{tabular}[c]{@{}l@{}}
Spreaders: 5,795 \\
Non-Spreaders: 18,545
\end{tabular}
\\
\hline
\end{tabular}
}
\end{table}

\paragraph{\textbf{Labeling Mechanisms:}} Due to the lack of ground-truth user-level annotations distinguishing misinformation spreaders from non-spreaders, prior research has predominantly relied on heuristic, threshold-based labeling strategies to construct supervised learning datasets.
We evaluate two complementary labeling strategies. 

The first, original threshold-based labeling follows CounterSpread \cite{82tommasel2025countering}; a user is labeled a spreader if more than 50\% of their shared tweets contain false claims. Under threshold-based labeling, user intent is ignored, which creates semantic noise in the labeling.

To address this, we proposed the second approach, stance-aware labeling, which is formally defined in Section \ref{Sec:Problemform}. Briefly, tweet-level stance predictions (support, deny, query, neutral) are generated using the GateNLP/stance-twitter-xlm-multilingual transformer. A user is labeled a spreader if their support ratio
$\phi(v,c_f)$ exceeds 0.5, with neutral tweets excluded. Formally
\begin{equation}
\phi(v, c_f) = |\text{support}(v)| \,/\,
(|\text{support}(v)| + |\text{deny}(v)| + |\text{query}(v)|) 
\end{equation}\\
The number of spreaders and non-spreaders obtained from both strategies is illustrated in Table \ref{tab:dataset_summary}.  Both strategies are evaluated using the described temporal settings.

\subsection{Temporal Modeling \& Evaluation Settings}
\label{subsec:tempeval}
SMSD is evaluated under two complementary experimental settings.
\begin{itemize}
    \item \textbf{Late-joining user prediction:} Users are chronologically divided into training, validation, and test sets in a 70/15/15 ratio based on their first-interaction timestamp. The model is trained on early users and evaluated on later joiners, with representations built from historical data available before each evaluation timestamp. 
    \item {\textbf{Early detection of emerging users:}} Two sub-settings are considered. 
    
    (i) Global Cumulative Split: Following CounterSpread \cite{82tommasel2025countering}, users are globally divided 70/30 by first-interaction timestamp. Tweets are partitioned into ten chronological quantile slots, Q1–Q10, with features reconstructed cumulatively up to each slot to prevent temporal leakage. 
    
    (ii) Monthly Quantile Split: The timeline is divided into twelve monthly slots, S = {Q1-Q12}. Within each month, the earliest 75\% of users are used to train the model, and the remaining 25\% form the test set. The monthly user split more accurately reflects how new user's join an ongoing misinformation event at different stages.
\end{itemize}

\paragraph{\textbf{Environmental Setup:}} The model was implemented in Python using PyTorch and PyTorch Geometric and evaluated with scikit-learn. Training was done on Google Colab Pro with an NVIDIA A100 GPU. 
\paragraph{\textbf{Research Questions: }}This study evaluates the performance of the SMSD approach by employing multiple labeling strategies, temporal evaluation protocols, and feature representations. The main research questions are as follows:
\begin{enumerate}
    \item RQ1: How does SMSD outperform the baselines when evaluated using various labeling strategies to predict the future behavior of late-joining users?
    \item RQ2: How early does SMSD detect emerging misinformation spreaders across different temporal settings and labeling strategies compared to baseline approaches?
\end{enumerate}

\subsection{Baselines \& Evaluation Metrics}
\label{subsec:baseeval}
We compared SMSD with several leading baseline methods, using different labeling strategies, temporal settings, and feature combinations to ensure a fair comparison. The baselines included traditional machine learning models, graph-based deep learning approaches, and recent misinformation spreader detection approaches such as CounterSpread \cite{82tommasel2025countering}, GraphSAGE \cite{124rath2022fake}, IHFS \cite{122chitti2024identifying}, and FakeThreads \cite{123patidar2025fakethreads}. 

Model performance is evaluated using Area Under the Curve (AUC), recall, precision, F1-score, weighted precision, weighted recall, weighted F1, and Matthews Correlation Coefficient (MCC). 

\begin{table}[t]
\setlength{\belowcaptionskip}{0pt}
\setlength{\abovecaptionskip}{0pt}
\caption{Performance comparison of SMSD with baseline methods under different labeling strategies (CS = CounterSpread, US = User Static, Beh = Behavioral Features, Bl = BLOC Features).}
\label{tab:rq1_main_comparison}
\centering
\renewcommand{\arraystretch}{1.2}
\setlength{\tabcolsep}{6pt} 
\resizebox{1.0\textwidth}{!}{
\begin{tabular}{|l|l|c|c|c|c|c|c|c|c|}
\hline

\textbf{Model} & \textbf{Features} & \textbf{AUC} & \textbf{Recall} & \textbf{Precision} & \textbf{F1} & \textbf{Prec$_w$} & \textbf{Rec$_w$} & \textbf{F1$_w$} & \textbf{MCC} \\
\hline

\multicolumn{10}{|c|}{\textbf{Original Threshold-Based Labeled Data}} \\
\hline

CS \cite{82tommasel2025countering}
& US
& 0.8701 & 0.7458 & 0.8427 & 0.7913 & 0.8101 & 0.8026 & 0.8101 & 0.6131 \\
\hline

SMSD
& US
& \textbf{0.9229} & \textbf{0.7763} & \textbf{0.8606} & \textbf{0.8163}
& \textbf{0.8545} & \textbf{0.8551} & \textbf{0.8539} & \textbf{0.6935} \\
\hline

CS \cite{82tommasel2025countering}
& US+Beh+Bl
& 0.8461 & \underline{0.7408} & 0.8192 & \underline{0.7825} & 0.8375 & 0.8381 & 0.8363 & \underline{0.6569} \\
\hline

SMSD
& US+Beh+Bl
& \textbf{0.9288} & \textbf{0.8718} & \textbf{0.8208} & \textbf{0.8402}
& \textbf{0.8726} & \textbf{0.8696} & \textbf{0.8704} & \textbf{0.7317} \\
\hline

\multicolumn{10}{|c|}{\textbf{Stance-aware Labeled Data}} \\
\hline

CS \cite{82tommasel2025countering}
& US+Beh+Bl
& \underline{0.7527} & \underline{0.5782} & 0.4382 & 0.5058 & 0.7493 & 0.7354 & \underline{0.7481} & \underline{0.3359} \\
\hline

GraphSAGE \cite{124rath2022fake}
& US+Beh+Bl
& 0.7469 & 0.4581 & \underline{0.4442} & 0.4510 & 0.7293 & 0.7266 & 0.7280 & 0.3030 \\
\hline

IHFS \cite{122chitti2024identifying}
& US+Beh+Bl
& 0.7240 & 0.5613 & 0.4193 & 0.5028 & 0.7435 & 0.7310 & 0.7364 & 0.3046 \\
\hline

FakeThreads \cite{123patidar2025fakethreads}
& US+Beh+Bl
& 0.7028 & 0.4908 & 0.4230 & \underline{0.5064} & \underline{0.7553} & \underline{0.7498} & 0.7474 & 0.3113 \\
\hline

SMSD
& US+Beh+Bl
& \textbf{0.8582} & \textbf{0.8386} & \textbf{0.6902} & \textbf{0.7572}
& \textbf{0.8078} & \textbf{0.7918} & \textbf{0.7944} & \textbf{0.5864} \\
\hline

\end{tabular}
}
\end{table}
\subsection{\textbf{Performance Analysis}} 
\label{subsec:perfanalysis}
\paragraph{\textbf{RQ1: Performance Comparison:}} Table~\ref{tab:rq1_main_comparison} presents a comparison of SMSD and baseline methods across two labeling strategies. With original threshold-based labeling, SMSD outperforms the best baseline (CounterSpread \cite{82tommasel2025countering} with US+Beh+Bl) by 17.7\% in recall, 7.4\% in F1-score, and 11.4\% in MCC. This indicates that behavioral and BLOC features provide additional discriminative information beyond profile, network, and conversation features. When using only static features, SMSD achieves an AUC of 92.29\% (versus 87.01\%) and an MCC that is 13.1\% higher than CounterSpread, demonstrating more effective user representation learning. Under stance-aware labeling, all baselines, including CounterSpread \cite{82tommasel2025countering}, IHFS \cite{122chitti2024identifying}, FakeThreads \cite{123patidar2025fakethreads}, and GraphSAGE \cite{124rath2022fake}, show significant performance drops due to increased labeling complexity and class imbalance. SMSD remains robust, with improvements of 45.0\% in recall over CounterSpread, 49.5\% in F1-score over FakeThreads, and 74.6\% in MCC over CounterSpread, as well as the highest AUC of 85.82\%.

\begin{table}[t]
\setlength{\belowcaptionskip}{0pt}
\setlength{\abovecaptionskip}{0pt}
\caption{Recall comparison under the Global Cumulative (GC) temporal split. (US = User Static, Beh = Behavioral Features, Bl = BLOC Features)}
\label{tab:rq2_gc}
\centering
\scriptsize
\renewcommand{\arraystretch}{1.2}
\setlength{\tabcolsep}{6pt} 

\resizebox{\textwidth}{!}{
\begin{tabular}{|l|l|c|c|c|c|c|c|c|c|c|c|}
\hline
\textbf{Model} &
\textbf{Features} &
\textbf{Q1} &
\textbf{Q2} &
\textbf{Q3} &
\textbf{Q4} &
\textbf{Q5} &
\textbf{Q6} &
\textbf{Q7} &
\textbf{Q8} &
\textbf{Q9} &
\textbf{Q10} \\
\hline

\multicolumn{12}{|c|}{\textbf{Original Threshold-Based Labeled Data}} \\
\hline

CS \cite{82tommasel2025countering}
& US
& 0.0000 & 0.0370 & 0.2703 & 0.0976
& 0.3881 & 0.3425 & 0.8190 & 0.8887
& 0.8805 & 0.8585 \\
\hline

SMSD
& US
& 0.2857 & 0.4037 & 0.5649 & 0.6049
& 0.7507 & 0.7630 & 0.8343 & 0.8987
& 0.9043 & 0.8674 \\
\hline

SMSD
& US+Beh+Bl
& \textbf{0.4286} & \textbf{0.7037}
& \textbf{0.7297} & \textbf{0.6829}
& \textbf{0.7612} & \textbf{0.9041}
& \textbf{0.9162} & \textbf{0.9022}
& \textbf{0.9028} & \textbf{0.9004} \\
\hline

\multicolumn{12}{|c|}{\textbf{Stance-Aware Labeled Data}} \\
\hline

CS \cite{82tommasel2025countering}
& US+Beh+Bl
& 0.5287 & 0.5199 & 0.5483 & 0.4960
& 0.4952 & 0.5402 & 0.5634 & 0.5353
& 0.5618 & 0.5437 \\
\hline

SMSD
& US+Beh+Bl
& \textbf{0.6204} & \textbf{0.5698}
& \textbf{0.5790} & \textbf{0.5392}
& \textbf{0.5373} & \textbf{0.5985}
& \textbf{0.6371} & \textbf{0.5561}
& \textbf{0.6500} & \textbf{0.6421} \\
\hline

\end{tabular}
}
\end{table}

\begin{table}[t]
\setlength{\belowcaptionskip}{0pt}
\setlength{\abovecaptionskip}{0pt}
\caption{Recall comparison under the Monthly Quantile (MQ) temporal split using stance-aware labels. (Here we are using User static, behavioral, and bloc features.)}
\label{tab:rq2_mq}
\centering
\scriptsize
\renewcommand{\arraystretch}{1.2}

\setlength{\tabcolsep}{3pt} 
\resizebox{1.0\textwidth}{!}{
\begin{tabular}{|l|c|c|c|c|c|c|c|c|c|c|c|c|}
\hline
\textbf{Model} &
\textbf{Q1} &
\textbf{Q2} &
\textbf{Q3} &
\textbf{Q4} &
\textbf{Q5} &
\textbf{Q6} &
\textbf{Q7} &
\textbf{Q8} &
\textbf{Q9} &
\textbf{Q10} &
\textbf{Q11} &
\textbf{Q12} \\
\hline

IHFS \cite{122chitti2024identifying}
& 0.2042 & 0.3249 & 0.3597 & 0.4072
& 0.4450 & 0.4633 & 0.4754 & 0.4819
& 0.4767 & \underline{0.4698} & 0.4844 & 0.4852 \\
\hline

FakeThreads \cite{123patidar2025fakethreads}

& 0.1028 & 0.2893 & 0.2910 & 0.3230
& 0.3560 & 0.3643 & 0.3783 & 0.3905
& 0.3812 & 0.3854 & 0.3948 & 0.3850 \\
\hline

GraphSAGE \cite{124rath2022fake}

& \underline{0.3132} & \underline{0.4772} & \underline{0.4851} & \underline{0.5448}
&\underline{0.4487} & \underline{0.5145} & \underline{0.4852} & \underline{0.5453}
& \underline{0.5045} & 0.4586 & \underline{0.5252} & \underline{0.4924} \\
\hline

SMSD

& \textbf{0.4680} & \textbf{0.5016}
& \textbf{0.5234} & \textbf{0.5676}
& \textbf{0.5816} & \textbf{0.6109}
& \textbf{0.6037} & \textbf{0.6353}
& \textbf{0.6419} & \textbf{0.6570}
& \textbf{0.6710} & \textbf{0.6879} \\
\hline

\end{tabular}
}
\end{table}

\paragraph{\textbf{RQ2: Performance Comparison:}} Tables~\ref{tab:rq2_gc} and~\ref{tab:rq2_mq} present temporal recall across all time slots. With threshold-based labeling, CounterSpread \cite{82tommasel2025countering} achieves 0.00\% recall in Q1 and 3.70\% in Q2, while SMSD with static features reaches 28.57\% in Q1. Adding behavioral and BLOC features further improves early detection, yielding 42.86\% recall in Q1, 70.37\% in Q2, and over 90.0\% from Q6 onward. These results show that temporal features support reliable early identification before sufficient interaction history is available. Under stance-aware labeling in the global cumulative setting, SMSD consistently outperforms the strongest baseline, CounterSpread \cite{82tommasel2025countering}, with a 17.4\% improvement in Q1 (62.04\% vs 52.87\%) and 15.7\% in Q9 (65.00\% vs 56.18\%).

In the monthly quantile setting, IHFS~\cite{122chitti2024identifying} and FakeThreads~\cite{123patidar2025fakethreads} plateau early, while GraphSAGE~\cite{124rath2022fake} fluctuates across slots. SMSD shows steady improvement from 46.80\% in Q1 to 68.79\% in Q12, outperforming GraphSAGE \cite{124rath2022fake} by 49.4\% in Q1 and 39.7\% in Q12. This demonstrates stable generalization to users joining at any stage of an ongoing misinformation event.

\section{Conclusion}
\label{sec:conc}
We introduce SMSD, a multimodal approach for detecting misinformation spreaders that integrates user profile, behavioral, BLOC, network-graph, and conversation-propagation features, along with a stance-aware labeling strategy and user-centric temporal evaluation. Evaluated on the FibVid COVID-19 dataset, SMSD consistently outperforms strong baselines across both labeling strategies and temporal settings. SMSD improves recall by 45.0\%, F1 by 49.5\%, and MCC by 74.6\% under stance-aware labeling and by 17.7\%, 7.4\%, and 11.4\%, respectively, under threshold-based labeling compared to the best competing method. Feature analysis shows that behavioral and BLOC features are essential for early detection, with BLOC alone delivering a $2.2{\times}$ improvement in recall over static features in Q1. Future work will validate SMSD on additional datasets, including AI-generated misinformation and other social media platforms. Incorporating large language models (LLMs) and transformer-based models may provide richer semantic and contextual representations.

%
\bibliographystyle{splncs04}
\bibliography{reference}

\end{document}